\documentclass[review,12pt]{elsarticle}
\usepackage{graphicx}
\usepackage{amsmath, amssymb}
\usepackage{booktabs}
\usepackage{hyperref}
\usepackage{subcaption}
\usepackage{float}
\usepackage{tikz}
\usepackage{pgfplots}
\pgfplotsset{compat=1.17}
\usepackage{caption}
\usepackage{tabularx} 
\usepackage{array} 
\usepackage{multirow}
\usepackage[most]{tcolorbox}
\usepackage{ltablex}
\usepackage{longtable}
\usepackage{adjustbox}

\journal{Neurocomputing}

\begin{document}

\begin{frontmatter}

\title{From Human Annotation to Automation: LLM-in-the-Loop Active Learning for Arabic Sentiment Analysis}

\author[1]{Dania Refai\corref{cor1}}
\author[1]{Alaa Dalaq}
\author[1]{Doaa Dalaq}
\author[1]{Irfan Ahmad}

\address[1]{Information and Computer Science Department, KFUPM, Dhahran 31261, Saudi Arabia}

\cortext[cor1]{Corresponding author: Dania Refai (E-mail: Dania.Refai@hotmail.com)}

\begin{abstract}
Natural language processing (NLP), particularly sentiment analysis, plays a vital role in areas like marketing, customer service, and social media monitoring by providing insights into user opinions and emotions. However, progress in Arabic sentiment analysis remains limited due to the lack of large, high-quality labeled datasets. While active learning has proven effective in reducing annotation efforts in other languages, few studies have explored it in Arabic sentiment tasks. Likewise, the use of large language models (LLMs) for assisting annotation and comparing their performance to human labeling is still largely unexplored in the Arabic context. In this paper, we propose an active learning framework for Arabic sentiment analysis designed to reduce annotation costs while maintaining high performance. We evaluate multiple deep learning architectures: Specifically, long short-term memory (LSTM), gated recurrent units (GRU), and recurrent neural networks (RNN), across three benchmark datasets: Hunger Station, AJGT, and MASAC, encompassing both modern standard Arabic and dialectal variations. Additionally, two annotation strategies are compared: Human labeling and LLM-assisted labeling. Five LLMs are evaluated as annotators: GPT-4o, Claude 3 Sonnet, Gemini 2.5 Pro, DeepSeek Chat, and LLaMA 3 70B Instruct. For each dataset, the best-performing LLM was used: GPT-4o for Hunger Station, Claude 3 Sonnet for AJGT, and DeepSeek Chat for MASAC. Our results show that LLM-assisted active learning achieves competitive or superior performance compared to human labeling. For example, on the Hunger Station dataset, the LSTM model achieved $93\%$ accuracy with only $450$ labeled samples using GPT-4o-generated labels, while on the MASAC dataset, DeepSeek Chat reached $82\%$ accuracy with $650$ labeled samples, matching the accuracy obtained through human labeling.
\end{abstract}


\end{frontmatter}

\section{Introduction}
Sentiment analysis is a fundamental task in natural language processing (NLP), aiming to determine the polarity of opinions provided in text: Positive, negative, or neutral. With the continuous growth of digital communication across social media, news platforms, and online forums, sentiment analysis has become an indispensable tool for understanding public opinion. Within this context, Arabic sentiment analysis is gaining increasing importance due to the rapid expansion of Arabic digital content and its socio-economic relevance across the Arab world~\cite{elsahar2015building}.

Despite this growing importance, progress in Arabic sentiment analysis remains limited compared to high-resource languages such as English, primarily due to the scarcity of large, high-quality labeled datasets~\cite{EDA}. The rich morphological complexity of Arabic and its diverse dialects further increases annotation costs and reduces the effectiveness of traditional supervised learning methods~\cite{zaghouani2014critical}.

To address these limitations, active learning has emerged as a promising paradigm. By iteratively selecting the most informative samples for annotation, active learning reduces labeling effort while maintaining model performance~\cite{settles2009active}, making it particularly well suited for resource-scarce settings such as Arabic.

Recent literature in Arabic sentiment analysis has explored curated datasets, lexicon-based methods, and traditional machine learning models such as support vector machine (SVM) and naive bayes (NB) \cite{b11,b20}. More recently, transformer-based approaches, including AraBERT and its variants, have achieved state-of-the-art results~\cite{b10,b18}, while aspect-based sentiment analysis and multi-task learning have further extended the scope of research~\cite{b15,b16}. Complementary to these efforts, active learning studies have proposed ensemble-based, uncertainty-driven, and voting-based strategies~\cite{6Anovel}, though only a few works have applied them in Arabic contexts, such as Moroccan dialect sentiment classification~\cite{7WeVoTe}. In parallel, the rise of large language models (LLMs) has introduced new opportunities, with frameworks like ActiveLLM and demonstration selection highlighting their potential in annotation and data acquisition~\cite{active1,active2}, alongside Bayesian, causal, and preference-based strategies~\cite{active3,active4,active8}.

Despite these advancements, important gaps remain at the intersection of active learning and Arabic sentiment analysis. In particular, active learning has seen limited adoption in Arabic contexts, and there is a lack of systematic comparisons between human annotation and LLM-assisted labeling within active learning frameworks.

In this paper, we address these gaps by introducing an active learning framework for Arabic sentiment analysis designed to reduce annotation costs while preserving high model performance. Our evaluation considers three deep learning architectures: Long short-term memory (LSTM), gated recurrent units (GRU), and recurrent neural networks (RNN), applied to three benchmark datasets (Hunger Station, AJGT, and MASAC), which together capture both modern standard Arabic and diverse dialectal variations. In addition, we examine the effectiveness of five LLMs: GPT-4o, Claude 3 Sonnet, Gemini 2.5 Pro, DeepSeek Chat, and LLaMA 3 70B Instruct, as annotation assistants, selecting the best-performing model for each dataset.

\subsection*{Main Contributions} Our main contributions can be summarized as follows:
\begin{itemize}  
    \item We propose an active learning framework for Arabic sentiment analysis that leverages large language models (GPT-4o, Claude 3 Sonnet, Gemini 2.5 Pro, DeepSeek Chat, and LLaMA 3 70B Instruct) as cost-efficient annotation assistants, demonstrating that annotation costs can be reduced without sacrificing model performance while identifying the most effective model for each dataset.   

    \item We conducted experiments on three benchmark datasets (Hunger Station, AJGT, MASAC) covering both modern standard Arabic and multiple dialects, ensuring that the evaluation reflects the linguistic diversity of real-world Arabic usage.

    \item We provide a systematic comparison between human annotations and LLM-generated labels, offering insights into the feasibility of LLMs as alternative annotators.  
\end{itemize}  

\subsection*{Paper Organization}The rest of the paper is organized as follows. In Section \ref{sec:literaturereview}, we present a comprehensive literature review. Our proposed methodology is explained in Section~\ref{sec:methodology}.  Section~\ref{sec:Experiments} discusses the experiments conducted to assess the robustness of the proposed approach along with their results and discussion. Finally, Section~\ref{sec:Conclusion} summarizes our conclusions and outlines potential directions for future research.


\section{Literature Review}\label{sec:literaturereview}

This section reviews recent research on active learning and significant contributions to Arabic sentiment analysis. The first part examines the application of active learning in both Arabic and non-Arabic contexts. The second part focuses on the state-of-the-art approaches in Arabic sentiment analysis, covering models, datasets, and domain-specific challenges. The discussion highlights key methodologies, learning strategies (supervised, weakly supervised, and active learning), feature representations, and sentiment granularity levels, including binary, multi-class, and aspect-based sentiment analysis.



\subsection{Active Learning}\label{subsec:active}

Active learning is a machine learning paradigm that aims to maximize model performance while minimizing the cost of data annotation. Unlike conventional supervised learning, which relies on large volumes of labeled data, active learning selectively queries the most informative samples for annotation, thereby reducing labeling requirements and associated expenses. This strategy is particularly valuable in domains where manual annotation is costly, time-consuming, or requires specialized expertise, as it enables the development of high-performing models with significantly fewer labeled instances \cite{ActiveLearning}. 

Salama \textit{et al.}~\cite{6Anovel}, introduce a novel enhanced weighted voting ensemble model (EWVAM) based on active learning. This model performs better than conventional methods, showcasing enhanced classification accuracy of $94.5\%$ and reduced training costs across various domains.
In another study by Matrane \textit{et al.}~\cite{7WeVoTe}, present an automatic annotation methodology designed explicitly for sentiment analysis within the Moroccan dialect. By leveraging advanced neural network models and employing techniques such as augmented stacking with weighted voting, the study achieved an accuracy of $95.18\%$ in sentiment analysis tasks for Moroccan dialect text.

Moreover, Bayer \textit{et al.}~\cite{active1} propose ActiveLLM, a new approach in which LLMs (such as GPT-4, LlaMA 3) score and select examples for few-shot learners, while the few-shot learners do not receive any in-loop training. They show that ActiveLLM mitigates the cold-start problem and results in a substantial improvement for BERT classifiers in few-shot scenarios. They achieved an F1-score of $0.7329$  using ActiveGPT4 with Prompt C2 (Class 2) on the CTI (Cyber threat intelligence) dataset. Margatina \textit{et al.}~\cite{active2} cast in-context demonstration selection as a one-shot, pool-based AL task over GPT and OPT models. They contrast uncertainty, diversity, and similarity sampling on 24 classification and multi-choice benchmarks, achieving an F1-score of $70\%$ using GPT-NeoX 20B on $15$ classification datasets. 
Additionally, Muldrew \textit{et al.}~\cite{active3} introduce active preference learning (APL) for direct preference optimization (DPO), and suggest a novel acquisition function that leverages predictive entropy and preference-model certainty. APL refines the model with DPO by repeatedly selecting prompt–completion pairs, querying their labels, and fine-tuning the model. They report an average accuracy of $76.1\%$ across 5 classification datasets.

Furthermore, Melo \textit{et al.}~\cite{active8} introduce BAL-PM, a Bayesian active learner to optimize epistemic uncertainty that maximizes the entropy of prompt distribution in the quickBATCH feature space. The batch nature of BAL-PM’s (Balanced active learning with prompt-based method) learning strategy also reduces redundancy and promotes diversity. Distill (Knowledge distillation) + TopK strategy on the MMLU (Massive multitask language understanding) benchmark achieved an average accuracy of $63.9\%$. In addition, a framework CAL (Causal-guided AL) developed by \textit{et al.} \cite{active4}, which makes use of LLMs to automatically detect when the dataset biases do not abide by causal invariance. CAL selects the most informative biased examples and induces explainable bias patterns, without the need for manual curation. They reported an average accuracy of $69.3\%$ on $14$ classification datasets using GPT-J (6B).

In contrast, Hassan \textit{et al.} \cite{active5} propose a clustering-based AL framework for generative tasks that incorporates clustering of the unlabeled outputs, in addition to LLM-guided knowledge distillation. Their loop loops between instance selection and refinement with a teacher LLM that has been trained with human-labeled distilled labels, achieving an accuracy of $63.9\%$ on the GSM8K (Math word problems) dataset using Active-Prompt-CoT.
On the other hand, Hübotter \textit{et al.} \cite{active6} present SIFT for test-time fine-tuning of LLMs, an acquisition algorithm that addresses the issue of information duplication and aims to achieve maximum information gain. SIFT (Scale-invariant feature transform) bridges retrieval and active learning, preventing redundant selection of nearest neighbors and prioritizing diverse relevant data. They reported an accuracy of $75.8\%$ using the FLAN-T5 model. Additionally, Xiao \textit{et al.} \cite{active7} introduce FreeAL, a human-free collaborative AL setting, where there is an LLM as an annotator and a small LRL (Low-resource language) model to filter the noisy labels. The LLM produces fuzzy in-context annotations, and the student SLM (Small language model) iteratively distills informative samples for label sharpening. They reported $0.557$ bits per byte (bpb) using LLaMA-3.2 (3B).


\subsection{Arabic Sentiment Analysis}\label{sec:ArabicSentAnalysis}

Arabic sentiment analysis focuses on detecting and interpreting opinions expressed in Arabic text, addressing unique linguistic challenges such as complex morphology, dialectal variations, and limited annotated resources. Research in this area has advanced through the development of specialized models, curated datasets, and preprocessing techniques tailored to the Arabic language \cite{arabic1}.

Mhamedi \textit{et al.}~\cite{b10} develop RMuBERT by pretraining, then fine-tuning, a pretrained transformer model on two newly collected semi-supervised news post datasets, (Arabic news posts) ANP5 and (Arabic news posts semi-supervised learning) ANPS2, which were built through selective loss functions and pseudolabeling. They compare Regional Multi-dialect BERT (RMuBERT) with Arabic BERT (AraBERT) and (Multi-dialect Arabic BERT) MARBERT, and obtain an accuracy of $95.43\%$ on AJGT (Arabic Jordanian general tweets) dataset~\cite{alomari2017ajgt} using AraBERTv2. Moreover, Ali and Manal~\cite{b11} apply the (National Research Council of Canada) NRC emotion lexicon with SVM, NB, and (K-nearest neighbor) KNN classifiers to $5,798$ Arabic tweets about online learning in COVID-19. Using an SVM, they achieved an accuracy of $89.6\%$ on the D2 dataset (Arabic tweets about online learning during the COVID-19 pandemic). Ishac \textit{et al.}~\cite{b20} study Arabic tweets for the 2022 FIFA world cup by leveraging their lexicon-based scoring classification models, SVM, and random forest, to analyze the temporal shifting trends of fandom sentiment. They construct a temporally matched collection of tweets related to a match and label them with polarity. Their method obtains more than $85\%$ accuracy in recognizing positive versus negative engagement concerning the key game events.

Alshaikh \textit{et al.} \cite{b15} introduce a BERT-based pipeline to process open-ended survey responses in Arabic, using attention to aspect embeddings to account for detailed feedback. They aggregate and annotate survey data from a customer satisfaction survey and further fine-tune BERT with an aspect-aware classification head. They achieve an F1-score of $0.86$ using FAST-LCF-ATEPC (Fine-grained Aspect-based sentiment classification with Transformers – local context focus – aspect term extraction and polarity classification) (multilingual) on task 2 (aspect polarity classification).

Additionally, Fadel \textit{et al.}~\cite{b16} present (Multi-task learning -AraBERT) MTL-AraBERT, a multi-task learning model, which simultaneously trains on the task of aspect extraction and sentiment classification on Arabic reviews based on shared encoder layers. They create a benchmark dataset across restaurants and e-commerce, and train their model with task-specific output heads. The AraBERT v0.2-large leads to a Macro-F1 Score of $0.9656$ on the aspect category sentiment classification (ACSC) Task.

Alosaimi \textit{et al.}~\cite{b18} inject ArabBERT contextual embeddings into an LSTM classifier (ArabBERT-LSTM) for generic social and news Arabic sentiment analysis. They use standard tokenization and fine-tune the integrated model on balanced and imbalanced corpora. They achieved an accuracy of $94.23\%$ on tweet sentiment classification.

Alotaibi \textit{et al.}~\cite{b27} present a weakly supervised paradigm with Snorkel to create labels from GPT-3.5, MarBERT, and XLM-RoBERTa (Cross-lingual model – RoBERTa) results of Arabic tweets related to Saudi educational reforms. They then fine-tune AraBERT on this weakly-labeled data, reaching $83\%$ precision, $76\%$ recall, and $85\%$ F1 in a three-way sentiment classification. Baniata and Kang~\cite{b35} proposed a switch-transformer model using a mixture-of-experts (MoE) encoder for the multitask sentiment analysis of Arabic dialects. They handle both five- and three-polarity tasks by separating input–output relations, through MoE and shared encoder layers. Tested on HARD (Hotel Arabic reviews dataset), BRAD (Book review Arabic dataset), and LABR (Large-scale Arabic Book reviews dataset) datasets, the model provides $84.02\%$, $67.89\%$, and $83.91\%$ accuracy, respectively. 

Alhumoud \textit{et al.}~\cite{b13} introduce ASAVACT, the biggest human-annotated Arabic COVID-19 vaccine tweet corpus ($32,476$ tweets) with seven expert annotations. They also compare SGRU (Stacked gated recurrent unit), Sbi-GRU (Stacked bidirectional gated recurrent unit), and an ensemble with AraBERT for pro vs. anti-vaccine classification. The MARBERT model achieved the best accuracy of $88.9\%$ on the ASTD (Arabic sentiment tweets dataset). More importantly, Mustafa \textit{et al.}~\cite{b14} investigate Food Delivery Service Reviews in the Arabic Language by employing CNN, BiLSTM, and LSTM-CNN hybrid models along with different stemmers and embeddings. They are trained on a hand-annotated Talabat dataset of modern standard Arabic and dialectal Arabic. Their performance reaches $84\%$ multiclass and $92.5\%$ binary classification accuracy for their best model. 

Mohammed and Kora~\cite{b22} collect a 40,000-tweet Arabic corpus and evaluate CNN, LSTM, and RCNN deep-learning models using word embeddings. The best accuracy is obtained by LSTM, 81.31\%, which can be improved to $89.61\%$ using data augmentation. In addition, Dandash and Asadpour~\cite{b23} present “AraPers” a dataset of $3,250$ Arabic Twitter users put together from users who posted 16Personalities test scores, and extract linguistic, profile, and metadata features. They apply varying degrees of models, from NB to BERT, and achieve an accuracy of $74.86\%$ for personality prediction. Khalil \textit{et al.}~\cite{b24} train a BiLSTM-based multi-label emotion analysis model, which is built on top of Aravec (Arabic word embeddings) and ARLSTEM (Arabic light stemmer) embeddings for Arabic tweets. They address the SemEval 2018 task 1 E-c multi-emotion classification task. Their model achieves the state-of-the-art F1 score of 96.6\%. 

Sabri \textit{et al.}~\cite{b25} propose ATCF-IDF, a term-weighting variant that is based on TF-IDF but incorporates the category-specific term frequency for Arabic text classification. They test ATCF-IDF with DT (Decision Tree), NB, and SVM on Khaleej-2004 and CNN-Arabic datasets. SVM with ATCF-IDF achieves $98.26\%$ accuracy, outperforming traditional weighting techniques. Jannani \textit{et al.}~\cite{b26} mix LDA (Latent dirichlet allocation) with Bi-GRU (Bidirectional gated recurrent unit) sentiment analysis of AraBERT and ASAFAYA embeddings on Moroccan news headlines. They collected likes and dislikes to tag sentiment and create a social well-being scoring and monitoring system. They reported state-of-the-art sentiment accuracy of $90.23\%$ and topic classification accuracy of $94.74\%$ using their Bi-GRU model with ASAFAYA.

An overview of the key models, datasets, and sentiment classification tasks reported in prior studies is presented in Table~\ref{tab:LR}. This compilation highlights the diversity of approaches employed in sentiment analysis, ranging from classical machine learning methods to advanced deep learning and transformer-based architectures. To provide a clearer conceptual structure, Fig.~\ref{fig:LR} illustrates a taxonomy of state-of-the-art techniques, categorizing them by learning strategy (Supervised, weakly supervised, and active learning), feature representation, and sentiment granularity (Binary, multi-class, and aspect-based analysis). This taxonomy serves as a reference framework for understanding the methodological landscape and the positioning of various approaches within the field.

\begin{table}[H]
\centering
\caption{Summary of existing studies on Arabic sentiment analysis and related tasks, including the models used, datasets, and corresponding tasks.}
\label{tab:LR}

\renewcommand{\arraystretch}{1.0}
\setlength{\tabcolsep}{2.5pt}
\scriptsize

\begin{adjustbox}{max width=\linewidth, max totalheight=\textheight, keepaspectratio}
\begin{tabular}{
  >{\centering\arraybackslash}p{1.2cm}   
  p{4.5cm}   
  p{4.5cm}   
  p{4.5cm}   
}
\toprule
\textbf{Authors} & \textbf{Model} & \textbf{Dataset} & \textbf{Task} \\
\midrule
\cite{b10} & RMuBERT & ANP5, ANPS2 & Sentiment analysis \\
\cite{b11} & NRC + ML classifiers & COVID-19 tweets & Sentiment analysis \\
\cite{b20} & Lexicon + SVM / Random forest & FIFA tweets & Sentiment tracking \\
\cite{b15} & BERT + Aspect attention & Survey responses & Aspect-aware sentiment analysis \\
\cite{b16} & MTL-AraBERT & Restaurant \& E-commerce reviews & Aspect \& Sentiment classification \\
\cite{b18} & ArabBERT-LSTM & Multiple Arabic corpora & Sentiment analysis \\
\cite{b27} & Snorkel + AraBERT & Saudi reform tweets & Sentiment classification \\
\cite{b35} & Switch-transformer + MoE & HARD, BRAD, LABR & Multi-task sentiment Analysis \\
\cite{b13} & Ensemble with AraBERT & ASAVACT & Vaccine Sentiment Analysis \\
\cite{b14} & CNN, BiLSTM, Hybrid & Talabat dataset & Food delivery sentiment \\
\cite{b21} & TF-IDF + Voting & Coffee tweets & Sentiment analysis \\
\cite{b22} & CNN, LSTM, RCNN & 40K Arabic tweets & Sentiment analysis \\
\cite{b23} & AraPers + ML / DL & 3,250 Personality tweets & Personality prediction \\
\cite{b24} & CNN-LSTM Hybrid & ATD dataset & Arabic sentiment classification \\
\cite{b25} & ATCF-IDF + ML & Khaleej-2004, CNN-Arabic & Text classification \\
\cite{b26} & LDA + Bi-GRU & Moroccan news & Sentiment \& Topic classification \\
\cite{6Anovel} & Ensemble + AL & Multiple sources & Classification \\
\cite{7WeVoTe} & Stacking + Voting & Moroccan dialect & Sentiment annotation \\
\cite{active1} & ActiveLLM & Few-shot benchmarks & Few-shot selection \\
\cite{active2} & Pool-based AL & 24 Benchmarks & In-context selection \\
\cite{active3} & APL + DPO & Preference datasets & Preference optimization \\
\cite{active8} & BAL-PM & Preference data & Prompt selection \\
\cite{active4} & CAL & Biased datasets & Bias detection \\
\cite{active5} & Clustering + LLM & Generative outputs & Generative selection \\
\cite{active6} & SIFT & Pile & Test-time tuning \\
\cite{active7} & FreeAL & LLM + SLM & Noisy label filtering \\
\bottomrule
\end{tabular}
\end{adjustbox}
\end{table}

\usetikzlibrary{positioning}

\begin{figure*}[ht]
\centering
\resizebox{\textwidth}{!}{%
\begin{tikzpicture}[
  every node/.style={font=\sffamily\Large},
  topnode/.style={
    rectangle, rounded corners, draw=orange!50!black,
    fill=yellow!20!white,
    align=center, text width=7.3cm, minimum height=1.9cm, thick
  },
  bottomnode/.style={
    draw,
    fill=green!10,
    rounded corners,
    text width=4.8cm,
    minimum height=7cm,   
    align=left,
    inner sep=6pt,        
  }
]

\node[topnode] (center) {\textbf{\large \mbox{Arabic Sentiment Classification}}};

\node[bottomnode, below=2.5cm of center, xshift=-10.5cm] (classical) {%
  \textbf{\large Classical Approaches}\\[2pt]
  \textbullet\ NRC + ML \cite{b11}\\
  \textbullet\ Lexicon + SVM/RF \cite{b20}\\
  \textbullet\ ATCF-IDF + ML \cite{b25}\\
  \textbullet\ TF-IDF + Voting \cite{b21}
};

\node[bottomnode, below=2.5cm of center, xshift=-5cm] (dl) {%
  \textbf{\large Deep Learning Models}\\[2pt]
  \textbullet\ CNN, LSTM, RCNN \cite{b22}\\
  \textbullet\ ArabBERT-LSTM \cite{b18}\\
  \textbullet\ CNN-LSTM Hybrid \cite{b24}\\
  \textbullet\ LDA + Bi-GRU \cite{b26}
};

\node[bottomnode, below=2.5cm of center, xshift=0.5cm] (transformer) {%
  \textbf{\large Transformer-based}\\[2pt]
  \textbullet\ RMuBERT \cite{b10}\\
  \textbullet\ BERT + Aspect Attention \cite{b15}\\
  \textbullet\ MTL-AraBERT \cite{b16}\\
  \textbullet\ AraBERT + Snorkel \cite{b27}\\
  \textbullet\ Switch Transformer + MoE \cite{b35}
};

\node[bottomnode, below=2.5cm of center, xshift=6.0cm] (al) {%
  \textbf{\large Active Learning (AL)}\\[2pt]
  \textbullet\ Ensemble + AL \cite{6Anovel}\\
  \textbullet\ FreeAL \cite{active7}\\
  \textbullet\ BAL-PM \cite{active8}\\
  \textbullet\ CAL \cite{active4}\\
  \textbullet\ SIFT \cite{active6}
};

\node[bottomnode, below=2.5cm of center, xshift=11.5cm] (weak) {%
  \textbf{\large Weak / Few-shot Learning}\\[2pt]
  \textbullet\ ActiveLLM \cite{active1}\\
  \textbullet\ Pool-based AL \cite{active2}\\
  \textbullet\ APL + DPO \cite{active3}\\
  \textbullet\ Clustering + LLM \cite{active5}
};

\foreach \target in {classical, dl, transformer, al, weak}
  \draw[->, thick] (center.south) -- (\target.north);

\end{tikzpicture}
}
\caption{Taxonomy of Arabic sentiment classification approaches.}
\label{fig:LR}
\end{figure*}
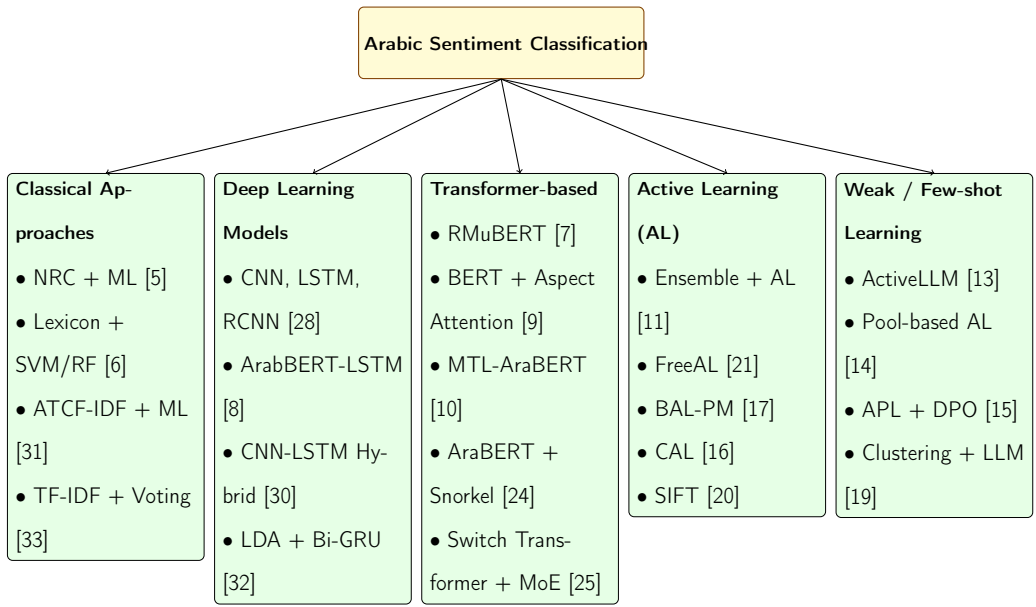

Although sentiment analysis and active learning have been extensively studied for high-resource languages such as English, research on Arabic remains limited. Existing studies suffer from three main shortcomings. First, there is a scarcity of labeled Arabic datasets covering diverse dialects, as most existing work focuses on modern standard Arabic or a single dialect, limiting model generalizability to real-world Arabic usage. Second, active learning, despite its potential to reduce annotation costs, has received little attention in the context of Arabic sentiment analysis. Third, to the best of our knowledge, no prior study has employed LLMs as annotators for Arabic sentiment analysis, and no work has systematically compared human annotations with LLM-generated annotations within an active learning framework, nor contrasted these results with scenarios that do not employ active learning. Addressing these gaps, this study investigates multiple deep learning models and a range of LLMs across diverse Arabic sentiment datasets to provide a comprehensive evaluation and advance the development of scalable, resource-efficient Arabic sentiment analysis systems.


\newpage
\section{Methodology}\label{sec:methodology}

This section presents a structured four-phase methodology for Arabic sentiment analysis that combines both non-active and active learning strategies. The primary objective of this framework is to reduce annotation effort through active learning while achieving performance levels comparable to those obtained with non-active learning approaches. Additionally, the methodology investigates the effectiveness of using LLMs as automated labelers compared to human annotators in the active learning setting. 

In the first phase, we analyze the characteristics of the selected sentiment datasets and apply essential preprocessing steps, including text normalization, tokenization, and cleaning. The second phase involves evaluating the predictive performance of five prominent LLMs: GPT-4o, Claude, Gemini, DeepSeek, and LLaMA on the preprocessed datasets. For each dataset, the model that achieves the highest predictive accuracy will be selected to carry out the labeling task in subsequent stages. The third phase focuses on training three deep learning models: RNN, LSTM, and GRU on fully labeled datasets to establish performance baselines. The fourth and final phase introduces active learning, wherein the model iteratively selects the most uncertain samples from an unlabeled data pool (Constructed based on the splits from phase two) for labeling and retraining. This cycle is repeated until the model reaches accuracy comparable to the baselines established in phase three. To evaluate labeling efficiency, we conduct two active learning scenarios: One using human-annotated labels and the other using labels generated by the best-performing LLM identified earlier in phase two. 

\begin{figure*}[h]
    \centering
    \includegraphics[width=1.0\linewidth]{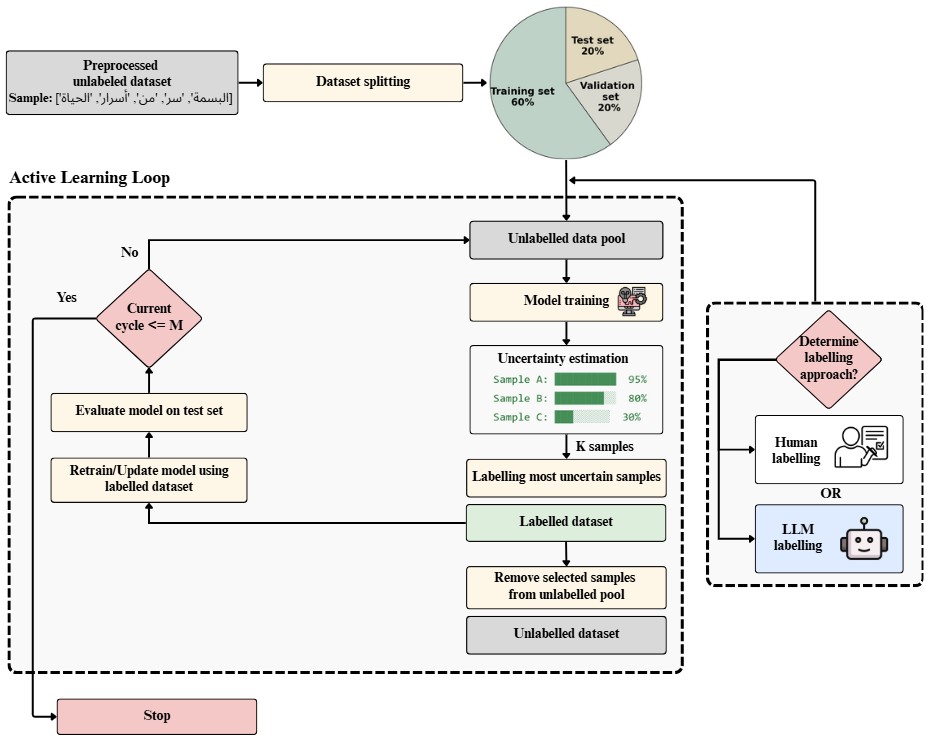}
    \caption{The proposed active learning framework for Arabic sentiment analysis begins with data preprocessing and splitting, followed by the selection of a labeling source, either an LLM or human annotators, before entering the active learning loop. Within the loop, the model is first trained, then iteratively selects the most uncertain samples for labeling by the chosen source, and retrains until a predefined threshold is met.}
    \label{fig:methodology}
\end{figure*}

An overview of the proposed active learning methodology is shown in Fig. \ref{fig:methodology}, and each phase is described in detail in the following sections.

\subsection{Phase 1: Datasets and Preprocessing}\label{phase1}

\subsubsection{Datasets}\label{datasets}

This study employs three distinct Arabic sentiment analysis datasets: A custom dataset collected from the Hunger Station mobile application~\cite{hungerstation2024} and two publicly available benchmark datasets, AJGT~\cite{alomari2017ajgt} and MASAC~\cite{bedi2021masac}.

The Hunger Station dataset, contains 51,409 user-written reviews in both modern standard Arabic and various local dialects. Although these reviews came with star ratings from 1 to 5, they did not include specific sentiment labels. To address this, we manually labeled a representative subset of 5,036 reviews based on their sentiment. This resulted in 2,399 positive, 2,506 negative, and 131 neutral reviews. The labeled data mainly includes reviews from the year 2023, although the full dataset spans five years. This annotated subset was then used for the experiments in this study.

To compare model performance and support generalization, we also used the well-known Arabic Jordanian general tweets (AJGT) dataset. This dataset includes 1,800 manually labeled tweets, split equally between positive and negative classes, with no neutral examples. The tweets are written in a mix of Jordanian dialect and modern standard Arabic and were labeled by native Arabic speakers to ensure accuracy. Because it comes from social media, the dataset reflects the informal, dialect-rich, and often noisy language found in real-life user content. This makes it a strong benchmark for testing sentiment models in realistic, low-resource environments.

Finally, the multi-domain Arabic sentiment corpus (MASAC) complements the other two datasets by providing a larger and more structured collection of Arabic texts. MASAC contains over 8,000 sentences from various domains, such as education, politics, health, and technology. Each sentence is labeled as either positive, negative, or neutral. Unlike AJGT, this dataset mostly uses formal modern standard Arabic and is grouped by topic, making it especially useful for testing models on more structured and grammatically correct texts from different fields.

All relevant details and statistics for the three datasets used in this study are summarized in Table~\ref{table:dataSets}.

\begin{table}[H]
\centering
\caption{Description of the Arabic sentiment analysis datasets considered for experimentation.}
\label{table:dataSets}
\renewcommand{\arraystretch}{1.2} 
\setlength{\tabcolsep}{4pt}      
\scriptsize                      

\begin{adjustbox}{max width=\linewidth}
\begin{tabular}{c l p{6.5cm} c l}
\toprule
\textbf{No.} & \textbf{Dataset} & \textbf{Description} & \textbf{Size} & \textbf{Classes} \\
\midrule
1 & Hunger Station \cite{hungerstation2024} & User reviews in Arabic from the HungerStation app, annotated for sentiment polarity and reflecting real-world opinions on services and products. & 4,500 & Negative, Neutral, Positive \\
2 & AJGT \cite{alomari2017ajgt} & Arabic tweets focused on the Jordanian dialect, labeled for binary sentiment classification (positive or negative). & 1,800 & Negative, Positive \\
3 & MASAC \cite{bedi2021masac} & Multi-dialectal Arabic corpus including both Modern Standard Arabic and dialects, annotated for three sentiment classes. Suitable for multiclass classification. & 2,000 & Negative, Neutral, Positive \\
\bottomrule
\end{tabular}
\end{adjustbox}
\end{table}


\subsubsection{Datasets Preprocessing}\label{DataPreprocessing}

To prepare the data for sentiment analysis, we applied a structured and unified preprocessing pipeline to all three datasets: Hunger Station (4,500 samples), AJGT (1,800 samples), and MASAC (2,000 samples). The goal was to clean the text, reduce noise, and handle the linguistic diversity of Arabic, ensuring that the data was consistent and meaningful for model training.

This pipeline included several steps, such as cleaning, normalization, tokenization, and linguistic filtering, and was applied uniformly across all datasets. Notably, we did not remove any samples during preprocessing. Instead, we focused on cleaning the internal content of each review or tweet. This ensured that the full set of annotated data was preserved and fully utilized throughout the experiments.

The detailed preprocessing steps are described below:

\begin{enumerate}
    \item \textbf{Removal of numerical characters:}  
    All numerical digits were removed from the samples. Since the study focuses solely on textual sentiment, numeric values were considered irrelevant and potentially misleading for the classification task.

    \item \textbf{Elimination of non-Arabic text and emojis:}  
    All non-Arabic characters, including Latin script, special symbols, and emojis, were filtered out. Emojis, while sometimes indicative of sentiment, were excluded in this study to maintain consistency and focus on linguistic signals within Arabic text. This step ensures that only native Arabic content is retained across all samples.

    \item \textbf{Punctuation removal:}  
    A comprehensive set of punctuation marks, including both general and Arabic-specific symbols, was removed to simplify the text and minimize syntactic noise.

    \item \textbf{Text normalization:}  
    Normalization was applied to reduce inconsistencies and standardize the text. Three specific sub-steps were performed:
    \begin{itemize}
        \item \textbf{Diacritics removal:} Arabic diacritical marks were removed, as they are often omitted in user-generated content and can introduce unnecessary variation.
        \item \textbf{Elongation (tatweel) removal:} Stylistic character extensions were normalized to their standard forms to reduce feature sparsity.
        \item \textbf{Emoji normalization and removal:} Emojis were not only removed but also normalized during earlier cleaning stages to ensure consistent treatment across the dataset.
    \end{itemize}
    Notably, we avoided normalization techniques that could alter word meaning or degrade the semantic integrity of the text.

    \item \textbf{Tokenization:}  
    The normalized text was tokenized into individual words using the Arabic tokenizer provided by the NLTK library. Tokenization enables further processing and feature extraction by structuring the text into discrete units.

    \item \textbf{Stop-word removal:}  
    High-frequency Arabic stop-words, were removed using NLTK’s built-in stop word list. These words typically carry low discriminative power for sentiment classification and can obscure relevant signals.

    \item \textbf{Stemming:}  
    To further reduce dimensionality and unify the morphological variants, stemming was performed using the Arabic stemmer from NLTK. This process transforms words to their base or root forms while preserving their core meaning.

    \item \textbf{Duplicate word removal:}  
    Repetitive words within the same review were eliminated to prevent artificially inflated term frequencies and reduce redundancy in the input space.
\end{enumerate} 


\subsection{Phase 2: Selecting the best LLM for each dataset}
\label{phase2}

In this phase, our objective is to identify the most effective LLM for each of the datasets introduced in Phase~\ref{phase1}. To accomplish this, we evaluated five state-of-the-art multilingual LLMs, accessed via the OpenRouter API, on their ability to perform sentiment classification in Arabic. For each dataset, we randomly sample $200$ Arabic reviews and provide them as input to each model. The models are instructed to predict a sentiment label for each review, and the predicted labels are then compared against the ground-truth annotations to compute classification accuracy. The model with the highest accuracy in each dataset is selected to serve as the automated labeling agent in the active learning process carried out in Phase~\ref{phase4}.

The evaluated models span both closed-source and open-source paradigms. Closed models include GPT-4o (OpenAI), Claude 3 Sonnet (Anthropic), and Gemini 2.5 Pro (Google), all of which are known for their strong performance on general-purpose multilingual tasks, including Arabic sentiment classification. On the open-source side, we evaluate DeepSeek Chat (DeepSeek) and LLaMA 3 $70$B Instruct (Meta), which have demonstrated competitive instruction follow-up capabilities and growing effectiveness in Arabic natural language processing.

To ensure consistent and deterministic results across all models, we configure the decoding parameters by setting the temperature to $0$ and limiting the maximum response length to $15$ tokens. Each model is prompted using a standardized instruction designed to constrain the output to a single predefined sentiment label, thereby avoiding variation in format and reducing ambiguity. The prompt is shown below:

\begin{tcolorbox}[
  colback=yellow!10!white,    
  colframe=brown!90!white,   
  boxrule=0.6pt,
  arc=2pt,
  sharp corners=south,
  title=\scriptsize\textbf{Sentiment Classification Prompt},
  fonttitle=\scriptsize\bfseries,
  width=0.7\linewidth, 
  enhanced,
  center               
]
\scriptsize
You will be given an Arabic review. Classify its sentiment as one of the following: \texttt{[LABELS]}.  
Respond with \textbf{ONLY ONE} label from this list. No explanation is needed.  

\textbf{Review:} \texttt{"..."}
\end{tcolorbox}

The key characteristics and configuration details of the evaluated models are summarized in Table~\ref{table:model_details}.

\begin{table}[H]
\centering
\caption{Summary of model details and configurations}
\label{table:model_details}
\renewcommand{\arraystretch}{1.1} 
\setlength{\tabcolsep}{4pt}      
\scriptsize                      

\begin{adjustbox}{max width=0.85\linewidth} 
\begin{tabular}{l c c c}
\toprule
\textbf{Model} & \textbf{Max Tokens} & \textbf{Source} & \textbf{Language} \\
\midrule
GPT-4o & 15 & Closed & Multilingual \\
Claude 3 Sonnet & 15 & Closed & Multilingual \\
Gemini 2.5 Pro & 15 & Closed & Multilingual \\
DeepSeek Chat & 15 & Open & Multilingual \\
LLaMA 3 70B Instruct & 15 & Open & Multilingual \\
\bottomrule
\end{tabular}
\end{adjustbox}
\end{table}

\subsection{Phase 3: Model Training with RNN, LSTM, and GRU Architectures}
\label{phase3}

This phase focuses on training and evaluating three deep learning models: RNN, LSTM, and GRU using the labeled dataset prepared in phase~\ref{phase1}. The datasets are divided into training, validation, and testing subsets with a $60\%$-$20\%$-$20\%$ split. Each model is trained until the validation loss stops improving for several consecutive epochs. The best performing model of each architecture is saved, and the corresponding test set scores are recorded for use in the subsequent active learning phase.

\subsubsection{Data preparation}

Prior to training, the labeled data is tokenized using the Tokenizer class with a vocabulary size limit of $2,000$ words. All sequences are padded to a fixed maximum length of $100$ tokens. Sentiment labels are numerically encoded through label encoding. The dataset is then partitioned into training ($60\%$), validation ($20\%$), and testing ($20\%$) sets.
\subsubsection{RNN model training}

Recurrent neural networks are designed to process sequential data by maintaining the memory of previous inputs. Their ability to retain context over time makes them suitable for tasks such as natural language processing and sentiment classification.

\textbf{Model architecture:} The RNN model is implemented using the TensorFlow Keras sequential API. It begins with an embedding layer configured with an input length of $100$, input dimension of $2,000$, and an output dimension of $32$, transforming text into dense vector representations. This is followed by two simpleRNN layers, each with $32$ units and dropout/recurrent dropout rates set to $0.2$ to mitigate overfitting. Batch normalization layers are added after each simpleRNN to stabilize and accelerate training. A final dense layer with a sigmoid activation function outputs the binary sentiment prediction.

\textbf{Training procedure:} The model is trained using the Adam optimizer and binary cross-entropy loss, with a batch size of $32$ over $20$ epochs. The performance of the model is monitored using the validation set, and the best model is selected based on the minimum validation loss. Table~\ref{tab:rnn_params} summarizes the RNN configuration.

\begin{table}[H]
\centering
\caption{Key parameters of the RNN model.}
\label{tab:rnn_params}
\renewcommand{\arraystretch}{1.2}
\setlength{\tabcolsep}{6pt}
\scriptsize

\begin{adjustbox}{max width=0.7\linewidth}
\begin{tabular}{p{5.0cm} p{3.0cm}}
\toprule
\textbf{Parameter} & \textbf{Value} \\
\midrule
Embedding input dim   & 2000 \\
Embedding output dim  & 32 \\
Embedding input length & 100 \\
SimpleRNN units       & 32 \\
Dropout rate          & 0.2 \\
Recurrent dropout rate & 0.2 \\
Activation function   & Sigmoid \\
Optimizer             & Adam \\
Loss function         & Binary Cross-Entropy \\
Number of epochs      & 20 \\
Batch size            & 32 \\
\bottomrule
\end{tabular}
\end{adjustbox}
\end{table}

\subsubsection{LSTM model training}

Long short-term memory networks are an advanced variant of RNNs, specifically designed to capture long-range dependencies through gated memory cells. Their architecture is well-suited for tasks involving sequential patterns, such as text classification.

\textbf{Model architecture:} The LSTM model is constructed using the sequential API. It includes an embedding layer with input dimension $2,000$, output dimension $32$, and input length $100$. This is followed by a single LSTM layer with 32 units and a dropout rate of $0.5$. A final dense layer with sigmoid activation is used for binary classification.

\textbf{Training procedure:} The model is trained using the Adam optimizer and binary cross-entropy loss for $20$ epochs with a batch size of $32$. Validation loss is used to track performance, and the best model is retained. Key configuration details are shown in Table~\ref{tab:lstm_params}.

\begin{table}[H]
\centering
\caption{Key parameters of the LSTM model.}
\label{tab:lstm_params}
\renewcommand{\arraystretch}{1.2}
\setlength{\tabcolsep}{6pt}
\scriptsize

\begin{adjustbox}{max width=0.7\linewidth}
\begin{tabular}{p{5.0cm} p{3.0cm}}
\toprule
\textbf{Parameter} & \textbf{Value} \\
\midrule
Input dimension      & 2000 \\
Output dimension     & 32 \\
Input length         & 100 \\
LSTM units           & 32 \\
Dropout rate         & 0.5 \\
Recurrent dropout    & 0.5 \\
Activation function  & Sigmoid \\
Optimizer            & Adam \\
Loss function        & Binary Cross-Entropy \\
Number of epochs     & 20 \\
Batch size           & 32 \\
\bottomrule
\end{tabular}
\end{adjustbox}
\end{table}

\subsubsection{GRU model training}

Gated recurrent units are a streamlined variant of LSTM networks designed to mitigate the problem of vanishing gradients in RNNs. They are capable of learning long-term dependencies while being computationally efficient.

\textbf{Model architecture:} The GRU model follows a similar sequential architecture, beginning with an embedding layer (input dimension of $2000$, output dimension of $32$, and input length of $100$). This is followed by a GRU layer with $16$ units and a dropout rate of $0.5$, after which batch normalization is applied. The final output layer is a dense layer with a sigmoid activation function, enabling binary classification.

\textbf{Training procedure:} The GRU model is trained using the Adam optimizer with binary cross-entropy loss over $100$ epochs and a batch size of $32$. Early stopping with a patience of $5$ epochs is applied to restore the best weights based on validation loss. The detailed GRU configuration is presented in Table~\ref{tab:gru_params}.

\subsubsection{Evaluation metrics}

To evaluate the performance of the RNN, LSTM, and GRU models, we calculate several metrics on the training, validation, and test sets, including accuracy, precision, recall, and F1-score. These metrics provide a comprehensive assessment of classification performance.

\begin{table}[H]
\centering
\caption{Key parameters of the GRU model.}
\label{tab:gru_params}
\renewcommand{\arraystretch}{1.2}
\setlength{\tabcolsep}{6pt}
\scriptsize

\begin{adjustbox}{max width=0.7\linewidth}
\begin{tabular}{p{5.0cm} p{3.0cm}}
\toprule
\textbf{Parameter} & \textbf{Value} \\
\midrule
Input dimension      & 2000 \\
Output dimension     & 32 \\
Input length         & 100 \\
GRU units            & 16 \\
Dropout rate         & 0.5 \\
Recurrent dropout    & 0.5 \\
Activation function  & Sigmoid \\
Optimizer            & Adam \\
Loss function        & Binary Cross-Entropy \\
Number of epochs     & 100 \\
Batch size           & 32 \\
\bottomrule
\end{tabular}
\end{adjustbox}
\end{table}


\subsection{Phase 4: Active Learning Framework}
\label{phase4}

In this phase, we implement an active learning framework to iteratively improve model performance by selectively labeling the most uncertain samples from an unlabeled data pool. The objective is to achieve test accuracy comparable to that obtained with non-active learning (As established in Phase~\ref{phase3}), while significantly reducing the number of manually labeled samples. This process is applied independently to each dataset.

To evaluate the efficiency of different annotation strategies, the active learning process is conducted twice for each dataset: Once using manual human labeling and once using automated labeling by the best performing LLM identified in Phase~\ref{phase2}. The results of both settings are compared based on test set accuracy and the number of cycles required to reach the target performance.

The active learning framework proceeds through the following steps:
\begin{enumerate}
    \item \textbf{Dataset initialization:}  
    For each dataset, the preprocessed data is divided into training, validation, and testing subsets. An \textit{unlabeled data pool} is then constructed from the training portion, consisting of samples without sentiment labels.

    \item \textbf{Active learning cycles:}  
    The model is iteratively trained through a sequence of active learning cycles. In each cycle, uncertain samples are selected for annotation, incorporated into the labeled set, and used for retraining. We set a threshold of $25$ active learning cycles, after which the labeling process stops. The test accuracy results across these cycles are then examined, and the labeling outcome corresponding to the first occurrence of test accuracy matching the best performance achieved without active learning in Phase~\ref{phase3} is selected. Each cycle involves the following stages:

    \begin{enumerate}
        \item \textbf{Model training:}  
        In the first cycle, the deep learning model that achieves the highest performance in the baseline (Non-active learning) phase is selected for use in the active learning experiments. This model is initially trained on a small set of 50 labeled samples. In subsequent cycles, it is retrained using the progressively expanded labeled set. Training is terminated when the validation loss fails to improve for several consecutive epochs.

        \item \textbf{Uncertainty estimation:}  
        An entropy-based uncertainty metric is applied to rank unlabeled samples according to prediction confidence, with priority given to those exhibiting the highest uncertainty.

        \item \textbf{Sample selection and labeling:}  
        In each cycle, a fixed number of the most uncertain samples (e.g., 50) are selected for annotation. For every dataset, the active learning process is conducted twice:
        \begin{itemize}
            \item \textit{Human annotation:} Samples are labeled by human annotators.
            \item \textit{LLM annotation:} Samples are labeled by the best-performing LLM from Phase~\ref{phase2}.
        \end{itemize}

        \item \textbf{Model update:}  
        The newly labeled samples are incorporated into the training data, and the model is retrained.

        \item \textbf{Performance evaluation:}  
        After each cycle, the model is evaluated on the fixed test set, and the test accuracy is recorded. Upon reaching the threshold of $25$ active learning cycles, the first cycle in which the model’s test accuracy matches the best accuracy obtained without active learning in Phase~\ref{phase3} is identified. The labeling results from that cycle are then selected for analysis.

    \end{enumerate}

\end{enumerate}
\section{Experimental Results and Discussion}
\label{sec:Experiments}

\subsection{Experimental Environment and Hardware} \label{EnvAndHard}
The experimental setup was implemented using Python $3$.$8$.$10$. All stages of the workflow,, including development, implementation, execution, and analysis, were carried out on a DELL Inspiron $15$ $5000$ laptop. This system is equipped with an Intel Core i$7$ processor, $16$GB of RAM, an NVIDIA GeForce MX$230$ GPU, and Intel Iris Plus integrated graphics. The machine runs on Windows $10$ Pro.

\subsection{LLMs Evaluation Results for Sentiment Labeling Across Different Datasets}
\label{sec:llm-results}
In this experiment, we evaluated the predictive performance of five LLMs: GPT-4o, Claude 3 Sonnet, Gemini 2.5 Pro, DeepSeek Chat, and LLaMA 3 70B Instruct on three Arabic sentiment datasets: HungerStation, AJGT, and MASAC. The primary objective was to determine, for each dataset individually, which model demonstrates the highest accuracy and should therefore be used to perform automated labeling during the active learning phase.

\begin{figure}[h]
    \centering
    \includegraphics[width=0.75\linewidth]{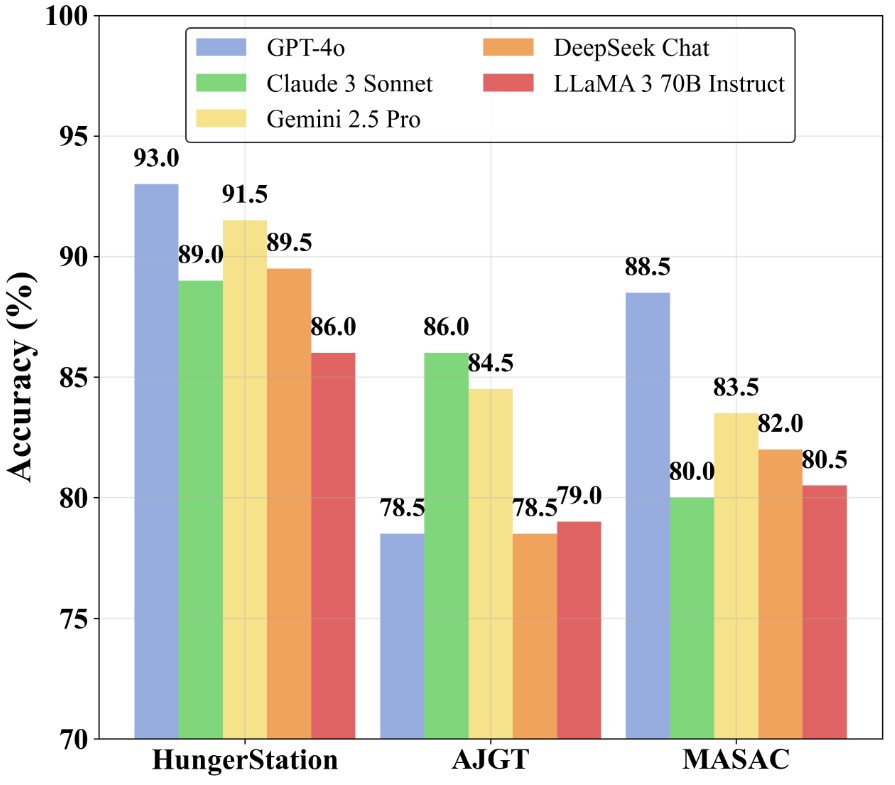}
    \caption{LLMs accuracy (\%) across all sentiment datasets}
    \label{fig:LLMsAccu}
\end{figure}

To ensure consistency and fairness in the evaluation, we randomly sampled $200$ reviews from each dataset and applied the same test samples across all models. Each model was queried using a standardized prompt format, which constrained the output to one of the predefined sentiment labels. Table~\ref{tab:llm-accuracy} and Fig.~\ref{fig:LLMsAccu} reports the accuracy of each model on all datasets.

\begin{table}[H]
\centering
\caption{Accuracy (\%) of LLMs across Arabic sentiment datasets.}
\label{tab:llm-accuracy}
\renewcommand{\arraystretch}{1.2}
\setlength{\tabcolsep}{10pt}
\scriptsize

\begin{adjustbox}{max width=0.8\linewidth}
\begin{tabular}{l c c c}
\toprule
\textbf{Model} & \textbf{HungerStation} & \textbf{AJGT} & \textbf{MASAC} \\
\midrule
GPT-4o              & \textbf{93.0} & 78.5 & \textbf{88.5} \\
Claude 3 Sonnet     & 89.0          & \textbf{86.0} & 80.0 \\
Gemini 2.5 Pro      & 91.5          & 84.5 & 83.5 \\
DeepSeek Chat       & 89.5          & 78.5 & 82.0 \\
LLaMA 3 70B Instruct& 86.0          & 79.0 & 80.5 \\
\bottomrule
\end{tabular}
\end{adjustbox}
\end{table}

The results reveal distinct differences in model performance across the sentiment datasets. Claude 3 Sonnet achieved the highest accuracy on the AJGT dataset, while GPT-4o outperformed all other models on both the HungerStation and MASAC datasets. Accordingly, we adopt a dataset-specific model selection strategy, whereby the model with the highest accuracy for a given dataset is employed for automated labeling in the active learning phase. This tailored approach ensures that the labeling process capitalizes on the strengths of the most effective model while accommodating the unique linguistic and contextual characteristics of each dataset.

\subsection{Performance of Deep Learning Models in the Non-Active Learning Setting}
\label{sec:nonactive-results}

To establish a baseline for subsequent experiments, we evaluated three deep learning architectures: LSTM, GRU, and RNN, on three Arabic sentiment analysis datasets: Hunger Station, AJGT, and MASAC. Each dataset is partitioned into $60$\% training, $20$\% validation, and $20$\% testing splits, resulting in $2,700$ samples for Hunger Station, $1,200$ for AJGT, and $1,080$ for MASAC.

The results, summarized in Table~\ref{table:combined_results}, represent the classification performance without incorporating any active learning strategies. For the Hunger Station dataset, LSTM attains the highest performance, marginally surpassing GRU and RNN. On the AJGT dataset, LSTM again achieves superior results, outperforming both GRU and RNN. A similar trend is observed for the MASAC dataset, where LSTM consistently yields the best performance.

Given its consistent superiority across all datasets, LSTM is selected as the benchmark model for the subsequent active learning phase, where its performance will be compared under various active learning strategies.

\begin{table*}[htbp]
\centering
\caption{Sentiment classification results without active learning across Hunger Station, AJGT, and MASAC datasets.}
\label{table:combined_results}
\renewcommand{\arraystretch}{1.2}
\scriptsize

\begin{tabular}{l l c c c c c}
\toprule
\textbf{Model} & \textbf{Dataset} & \textbf{\# Samples} & \textbf{Accuracy} & \textbf{Precision} & \textbf{Recall} & \textbf{F1 Score} \\
\midrule
\multirow{3}{*}{LSTM} 
& Hunger Station & 2700 & 0.93 & 0.93 & 0.87 & 0.90 \\
& AJGT           & 1200 & 0.83 & 0.85 & 0.83 & 0.84 \\
& MASAC          & 1080 & 0.82 & 0.82 & 0.81 & 0.82 \\
\midrule
\multirow{3}{*}{RNN} 
& Hunger Station & 2700 & 0.92 & 0.89 & 0.87 & 0.88 \\
& AJGT           & 1200 & 0.52 & 0.55 & 0.58 & 0.56 \\
& MASAC          & 1080 & 0.50 & 0.61 & 0.50 & 0.55 \\
\midrule
\multirow{3}{*}{GRU} 
& Hunger Station & 2700 & 0.92 & 0.91 & 0.87 & 0.89 \\
& AJGT           & 1200 & 0.81 & 0.88 & 0.76 & 0.81 \\
& MASAC          & 1080 & 0.75 & 0.82 & 0.72 & 0.77 \\
\bottomrule
\end{tabular}
\end{table*}

\subsection{Active Learning Performance Evaluation}

This section presents the evaluation of the active learning framework applied to the LSTM model across the HungerStation, AJGT, and MASAC datasets. The LSTM architecture was selected for this phase based on its consistently superior performance in the baseline (Non-active learning) experiments. Two independent experiments are conducted: one using human-labeled data and the other using LLM-generated labels. The following subsections report the results of each experiment separately, covering active learning with human-labeled data and active learning with LLM-labeled data.


\subsubsection{Active Learning Using Human-Labeled Data}

In this experiment, active learning is applied to all three datasets: Hunger Station, AJGT, and MASAC, using the LSTM model. In each cycle, a human annotator labels the newly selected samples, which are then added to the labeled set for model retraining. This process is repeated until all active learning cycles are completed.

For the Hunger Station dataset (Table~\ref{table:HS-human}), the model’s accuracy improves substantially as labeled samples are progressively added. While the first cycle begins with an accuracy of $0.35$, performance increases steadily thereafter. By the ninth cycle, accuracy reaches $0.87$, and by the fourteenth cycle, with only $700$ labeled samples, it attains $0.93$, matching the baseline accuracy of $0.93$ achieved in the non-active learning setting using the full $2,700$ labeled samples. This shows that active learning can achieve baseline accuracy with far fewer labeled samples.

\begin{table}[H]
\centering
\caption{Sentiment classification results for LSTM model with active learning on the Hunger Station dataset (Human labeling).}
\label{table:HS-human}
\scriptsize
\renewcommand{\arraystretch}{0.8}
\setlength{\tabcolsep}{6pt} 

\begin{tabular}{c c c c c c}
\toprule
\textbf{Cycle} & \textbf{Label Count} & \textbf{Accuracy} & \textbf{Precision} & \textbf{Recall} & \textbf{F1 Score} \\
\midrule
1  & 50  & 0.35 & 0.34 & 0.77 & 0.51 \\
2  & 100 & 0.59 & 0.45 & 0.67 & 0.33 \\
3  & 150 & 0.34 & 0.34 & 0.77 & 0.51 \\
4  & 200 & 0.55 & 0.23 & 0.60 & 0.55 \\
5  & 250 & 0.35 & 0.35 & 0.77 & 0.51 \\
6  & 300 & 0.52 & 0.44 & 0.55 & 0.50 \\
7  & 350 & 0.57 & 0.42 & 0.58 & 0.48 \\
8  & 400 & 0.80 & 0.68 & 0.81 & 0.74 \\
9  & 450 & 0.87 & 0.82 & 0.83 & 0.82 \\
10 & 500 & 0.89 & 0.84 & 0.87 & 0.86 \\
11 & 550 & 0.91 & 0.88 & 0.87 & 0.88 \\
12 & 600 & 0.91 & 0.87 & 0.88 & 0.87 \\
13 & 650 & 0.92 & 0.89 & 0.88 & 0.89 \\
14 & 700 & 0.93 & 0.92 & 0.87 & 0.89 \\
\bottomrule
\end{tabular}
\end{table}

On the AJGT dataset (Table~\ref{table:AJGT-human}), the model shows consistent improvement as more labeled data is added. Beginning with only $50$ labeled instances, it achieves an initial accuracy of $0.54$, which rises steadily across cycles. By the ninth cycle ($450$ samples), accuracy reaches $0.77$, and by the final cycle with $700$ labeled samples, it attains $0.83$, closely matching the baseline accuracy of $0.83$ obtained in the non-active learning setting with the full $1,200$ labeled samples. These results confirm the efficiency of active learning in achieving competitive accuracy with significantly fewer annotations.

\begin{table}[H]
\centering
\caption{Sentiment classification results for the LSTM model with active learning on the AJGT dataset (Human labeling).}
\label{table:AJGT-human}
\scriptsize
\renewcommand{\arraystretch}{0.8}
\setlength{\tabcolsep}{6pt} 

\begin{tabular}{c c c c c c}
\toprule
\textbf{Cycle} & \textbf{Label Count} & \textbf{Accuracy} & \textbf{Precision} & \textbf{Recall} & \textbf{F1 Score} \\
\midrule
1  & 50  & 0.54 & 0.54 & 0.98 & 0.67 \\
2  & 100 & 0.54 & 0.54 & 1.00 & 0.70 \\
3  & 150 & 0.54 & 0.54 & 1.00 & 0.69 \\
4  & 200 & 0.61 & 0.60 & 0.90 & 0.71 \\
5  & 250 & 0.67 & 0.65 & 0.83 & 0.73 \\
6  & 300 & 0.68 & 0.65 & 0.87 & 0.74 \\
7  & 350 & 0.72 & 0.70 & 0.88 & 0.77 \\
8  & 400 & 0.74 & 0.73 & 0.81 & 0.77 \\
9  & 450 & 0.77 & 0.75 & 0.86 & 0.80 \\
10 & 500 & 0.78 & 0.78 & 0.83 & 0.80 \\
11 & 550 & 0.79 & 0.81 & 0.80 & 0.80 \\
12 & 600 & 0.81 & 0.81 & 0.84 & 0.83 \\
13 & 650 & 0.81 & 0.86 & 0.78 & 0.82 \\
14 & 700 & 0.83 & 0.84 & 0.84 & 0.84 \\
\bottomrule
\end{tabular}
\end{table}

For the MASAC dataset (Table~\ref{table:MASAC-human}), the model starts with an accuracy of $0.49$ in the early cycles. As additional labeled samples are incorporated, accuracy improves steadily, reaching $0.82$ by the fifteenth cycle with only $750$ labeled instances. This matches the baseline accuracy of $0.82$ achieved in the non-active learning setting using the full $1,080$ labeled samples. These results demonstrate that active learning can reach baseline-level accuracy with considerably fewer annotations.

\begin{table}[H]
\centering
\caption{Sentiment classification results for the LSTM model with active learning on the MASAC dataset (Human labeling).}
\label{table:MASAC-human}
\scriptsize
\renewcommand{\arraystretch}{0.8}
\setlength{\tabcolsep}{6pt} 

\begin{tabular}{c c c c c c}
\toprule
\textbf{Cycle} & \textbf{Label Count} & \textbf{Accuracy} & \textbf{Precision} & \textbf{Recall} & \textbf{F1 Score} \\
\midrule
1   & 50  & 0.49 & 0.49 & 1.00 & 0.66 \\
2   & 100 & 0.49 & 0.49 & 1.00 & 0.66 \\
3   & 150 & 0.49 & 0.49 & 1.00 & 0.66 \\
4   & 200 & 0.52 & 0.50 & 0.95 & 0.66 \\
5   & 250 & 0.58 & 0.54 & 0.94 & 0.69 \\
6   & 300 & 0.57 & 0.54 & 0.95 & 0.69 \\
7   & 350 & 0.58 & 0.54 & 0.93 & 0.68 \\
8   & 400 & 0.65 & 0.60 & 0.86 & 0.71 \\
9   & 450 & 0.70 & 0.65 & 0.84 & 0.73 \\
10  & 500 & 0.74 & 0.70 & 0.80 & 0.75 \\
11  & 550 & 0.73 & 0.71 & 0.77 & 0.74 \\
12  & 600 & 0.77 & 0.75 & 0.80 & 0.77 \\
13  & 650 & 0.81 & 0.80 & 0.81 & 0.81 \\
14  & 700 & 0.81 & 0.79 & 0.84 & 0.82 \\
15  & 750 & 0.82 & 0.80 & 0.84 & 0.82 \\
\bottomrule
\end{tabular}
\end{table}

Overall, across all three datasets, active learning with human annotators achieves accuracy equivalent to that obtained in the baseline (Non-active learning) setting while requiring substantially fewer labeled samples. This confirms its effectiveness in reducing annotation demands without compromising model performance.

\subsubsection{Active Learning Using LLM-Labeled Data}

In this experiment, active learning is applied to the Hunger Station, AJGT, and MASAC datasets using the LSTM model, with annotations provided by the best-performing LLM selected for each dataset in the previous experiment. In each cycle, the newly selected samples are labeled by the corresponding LLM and added to the training set for model retraining. This process continues until all active learning cycles are completed.

For the Hunger Station dataset (Table~\ref{table:HS-LLM}), the model begins with an accuracy of $0.78$ in the first cycle and improves steadily, reaching $0.90$ by the eighth cycle with only $400$ labeled samples. This performance is maintained in the final cycle ($450$ samples), which is close to the baseline non-active learning accuracy of $0.93$ obtained with $2,700$ labeled samples.

\begin{table}[H]
\centering
\caption{Sentiment classification results for the LSTM model with active learning on the Hunger Station dataset (LLM labeling).}
\label{table:HS-LLM}
\scriptsize
\renewcommand{\arraystretch}{0.8}
\setlength{\tabcolsep}{6pt} 

\begin{tabular}{c c c c c c}
\toprule
\textbf{Cycle} & \textbf{Label Count} & \textbf{Accuracy} & \textbf{Precision} & \textbf{Recall} & \textbf{F1 Score} \\
\midrule
1  & 50  & 0.78 & 0.64 & 0.83 & 0.72 \\
2  & 100 & 0.86 & 0.80 & 0.78 & 0.79 \\
3  & 150 & 0.74 & 0.95 & 0.26 & 0.41 \\
4  & 200 & 0.88 & 0.86 & 0.77 & 0.81 \\
5  & 250 & 0.87 & 0.87 & 0.73 & 0.80 \\
6  & 300 & 0.88 & 0.95 & 0.68 & 0.79 \\
7  & 350 & 0.89 & 0.92 & 0.75 & 0.83 \\
8  & 400 & 0.90 & 0.91 & 0.80 & 0.85 \\
9  & 450 & 0.93 & 0.92 & 0.80 & 0.85 \\
\bottomrule
\end{tabular}
\end{table}

On the AJGT dataset (Table~\ref{table:AJGT-LLM}), the model starts with an accuracy of $0.54$ using $50$ labeled samples. Accuracy increases gradually across cycles, reaching $0.73$ by the fourth cycle ($200$ samples) and stabilizing at $0.75$ in the final cycle ($550$ samples). Although this remains below the baseline accuracy of $0.83$ achieved with $1,200$ labeled samples, it still demonstrates meaningful gains with fewer annotations.

\begin{table}[H]
\centering
\caption{Sentiment classification results for the LSTM model with active learning on the AJGT dataset (LLM labeling).}
\label{table:AJGT-LLM}
\scriptsize
\renewcommand{\arraystretch}{0.8}
\setlength{\tabcolsep}{6pt} 

\begin{tabular}{c c c c c c}
\toprule
\textbf{Cycle} & \textbf{Label Count} & \textbf{Accuracy} & \textbf{Precision} & \textbf{Recall} & \textbf{F1 Score} \\
\midrule
1  & 50  & 0.54 & 0.54 & 1.00 & 0.70 \\
2  & 100 & 0.64 & 0.72 & 0.55 & 0.62 \\
3  & 150 & 0.59 & 0.71 & 0.39 & 0.51 \\
4  & 200 & 0.73 & 0.79 & 0.69 & 0.74 \\
5  & 250 & 0.71 & 0.82 & 0.59 & 0.69 \\
6  & 300 & 0.70 & 0.74 & 0.67 & 0.70 \\
7  & 350 & 0.72 & 0.84 & 0.61 & 0.71 \\
8  & 400 & 0.74 & 0.80 & 0.59 & 0.69 \\
9  & 450 & 0.74 & 0.86 & 0.61 & 0.72 \\
10 & 500 & 0.72 & 0.83 & 0.60 & 0.69 \\
11 & 550 & 0.75 & 0.84 & 0.65 & 0.73 \\
\bottomrule
\end{tabular}
\end{table}

For the MASAC dataset (Table~\ref{table:MASAC-LLM}), the model begins with an accuracy of $0.59$ in the first cycle and reaches $0.71$ by the fifth cycle ($250$ samples). By the final cycle ($650$ samples), it attains $0.82$, matching the baseline accuracy of $0.82$ obtained with $1,080$ labeled samples in the non-active learning setting.

\begin{table}[H]
\centering
\caption{Sentiment classification results for the LSTM model with active learning on the MASAC dataset (LLM labeling).}
\label{table:MASAC-LLM}
\scriptsize
\renewcommand{\arraystretch}{0.8}
\setlength{\tabcolsep}{6pt} 

\begin{tabular}{c c c c c c}
\toprule
\textbf{Cycle} & \textbf{Label Count} & \textbf{Accuracy} & \textbf{Precision} & \textbf{Recall} & \textbf{F1 Score} \\
\midrule
1   & 50  & 0.59 & 0.81 & 0.25 & 0.38 \\
2   & 100 & 0.66 & 0.60 & 0.75 & 0.69 \\
3   & 150 & 0.70 & 0.72 & 0.64 & 0.68 \\
4   & 200 & 0.69 & 0.73 & 0.61 & 0.65 \\
5   & 250 & 0.71 & 0.72 & 0.67 & 0.70 \\
6   & 300 & 0.73 & 0.76 & 0.66 & 0.71 \\
7   & 350 & 0.74 & 0.75 & 0.69 & 0.71 \\
8   & 400 & 0.74 & 0.74 & 0.63 & 0.70 \\
9   & 450 & 0.75 & 0.82 & 0.64 & 0.72 \\
10  & 500 & 0.75 & 0.76 & 0.71 & 0.74 \\
11  & 550 & 0.75 & 0.82 & 0.68 & 0.74 \\
12  & 600 & 0.75 & 0.80 & 0.66 & 0.72 \\
13  & 650 & 0.82 & 0.80 & 0.73 & 0.76 \\
\bottomrule
\end{tabular}
\end{table}

Overall, across all three datasets, active learning with LLM-generated labels achieves accuracy comparable to the baseline (Non-active learning) results while requiring substantially fewer labeled samples. These findings demonstrate the potential of LLM-based annotation to reduce manual labeling costs without compromising model performance.

\subsection{Overall Findings and Comparative Analysis}

This study provides a comprehensive evaluation of active learning applied to sentiment classification across different datasets, while comparing human annotation with LLM-based labeling. The analysis quantifies the reduction in annotation effort enabled by active learning and assesses the viability of LLM-generated labels as an alternative to traditional human annotation.  

Our findings establish two central contributions. First, we demonstrate that active learning can consistently achieve baseline-level performance, comparable to that obtained with fully labeled datasets, while requiring only a fraction of the labeling effort. This performance consistency holds across datasets and annotation sources, underscoring active learning’s robustness and efficiency in reducing the volume of labeled data without sacrificing accuracy. Second, we show that LLM-generated labels, when integrated into active learning pipelines, can closely match the effectiveness of human annotation, in some cases reaching equivalent accuracy in fewer labeling cycles.

\begin{figure}[H]
\centering

\begin{subfigure}{\linewidth}
  \centering
  \includegraphics[width=0.62\linewidth]{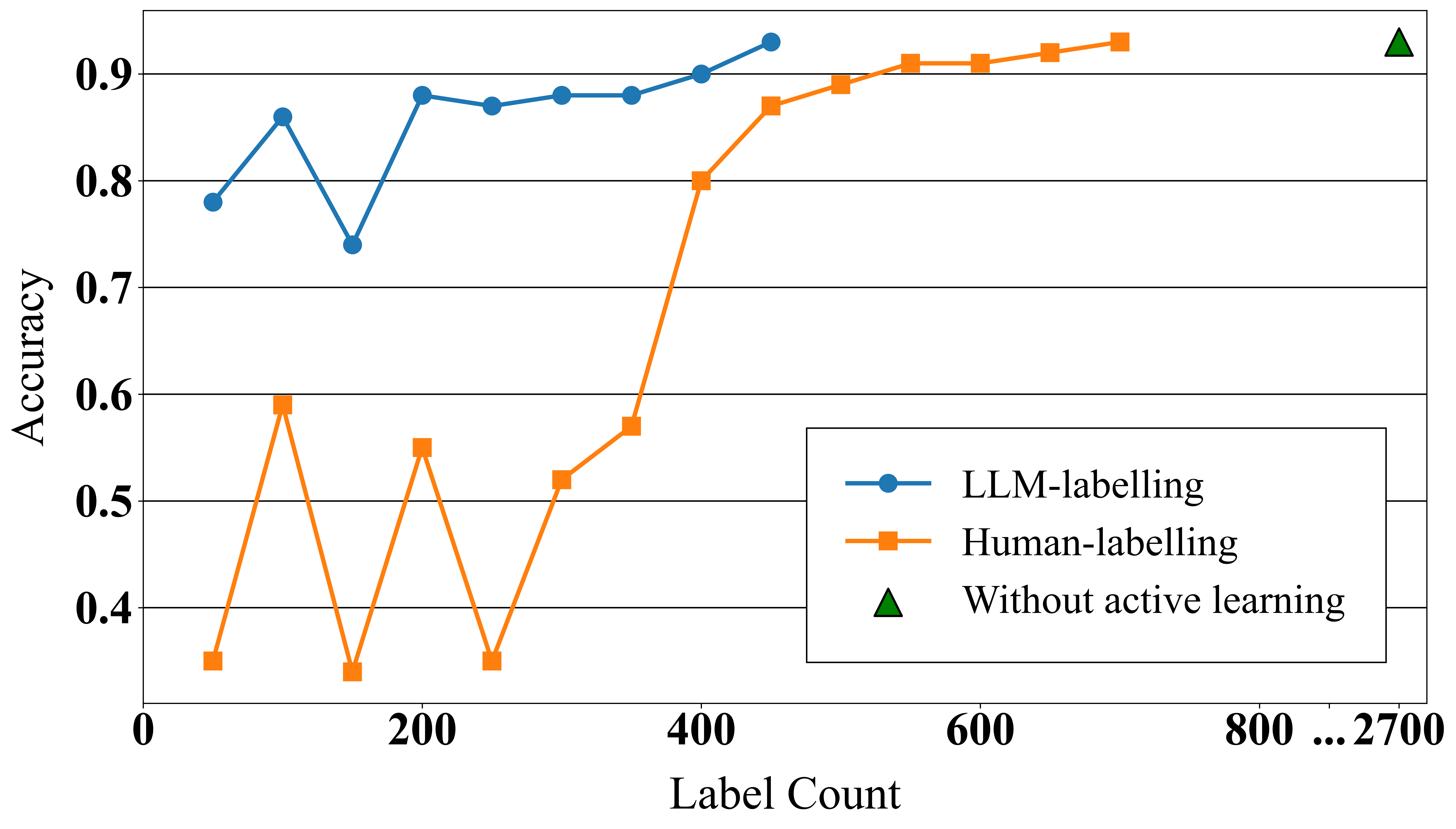}
  \caption{Hunger Station dataset.}
\end{subfigure}

\vspace{0.5em}

\begin{subfigure}{\linewidth}
  \centering
  \includegraphics[width=0.62\linewidth]{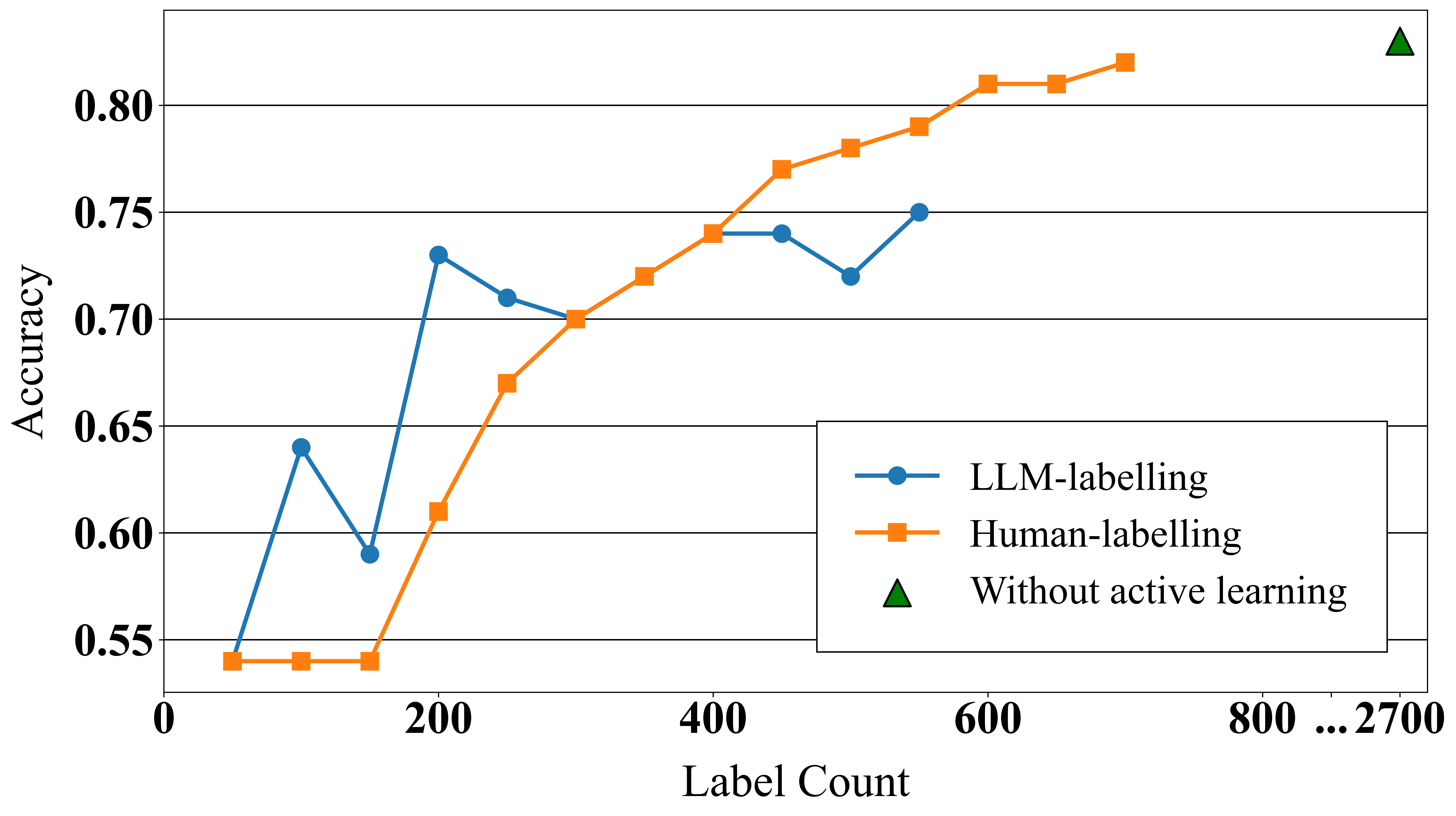}
  \caption{AJGT dataset.}
\end{subfigure}

\vspace{0.5em}

\begin{subfigure}{\linewidth}
  \centering
  \includegraphics[width=0.62\linewidth]{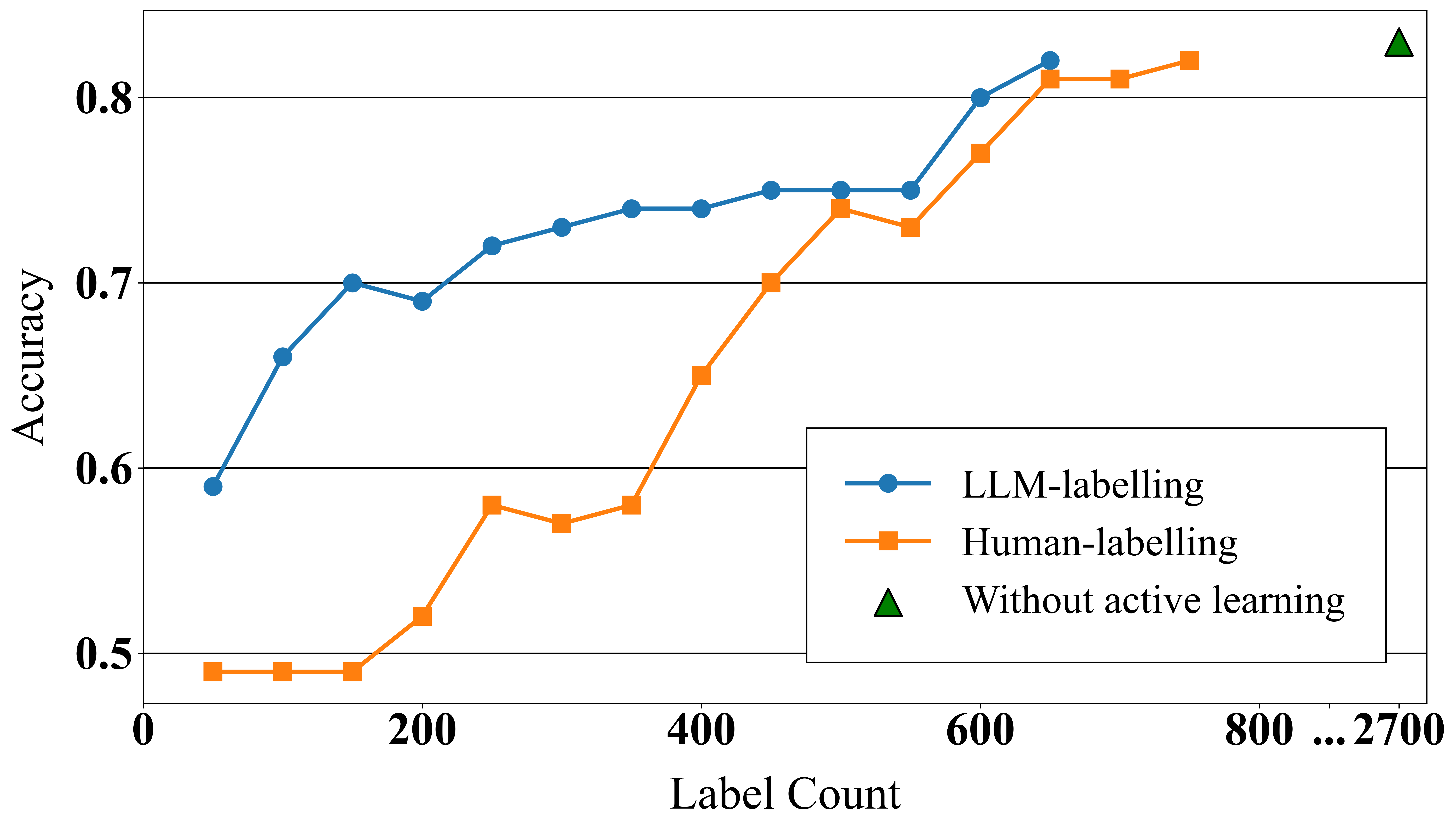}
  \caption{MASAC dataset.}
\end{subfigure}

\caption{Accuracy comparison across Hunger Station, AJGT, and MASAC datasets under different labeling strategies. LLM-assisted labeling achieves higher accuracy with fewer samples than human labeling and consistently outperforms the non-active learning baseline.}
\label{fig:exp3results}
\end{figure}

The performance trends are illustrated in Fig.~\ref{fig:exp3results}. For the Hunger Station dataset, human annotation reached the baseline accuracy of $0.93$ after $14$ cycles with $700$ labeled samples, while LLM labeling achieved the same performance in just $9$ cycles with $450$ samples. Early cycles by human annotation fluctuated heavily between $0.4$–$0.6$ accuracy, whereas LLMs maintained stable performance above $0.8$ throughout, showing clear efficiency gains. In the AJGT dataset, human annotation ultimately outperformed LLMs, reaching $0.83$ in $14$ cycles ($700$ samples), compared to $0.75$ with $11$ cycles ($550$ samples) for LLMs. However, LLM-assisted labeling provided faster early improvements, highlighting its sample efficiency even if humans achieved slightly higher final accuracy. For the MASAC dataset, both strategies converged to comparable performance ($0.82$), but LLM labeling reached this level with $650$ samples versus $750$ for human annotation, and its accuracy curve showed smoother, more consistent gains.

Overall, this research contributes in two significant ways. First, it empirically confirms that active learning can substantially reduce the amount of labeled data required to achieve high classification accuracy, thereby enhancing the efficiency of sentiment analysis. Second, it shows that LLMs can serve as an effective replacement for human annotation when combined with active learning. This approach accelerates the labeling process, optimizes resource allocation, and enables the development of more scalable sentiment analysis systems.




\section{Conclusions and Future Research}
\label{sec:Conclusion}
Arabic sentiment analysis is increasingly important for understanding public opinion on digital platforms. However, its advancement is hindered by the limited availability of annotated datasets and the linguistic complexity of Arabic, particularly its diverse dialects and rich morphology. While manual annotation is accurate, it is both time-consuming and resource-intensive, posing scalability challenges. Active learning emerges as a promising alternative by strategically minimizing labeling effort, yet its application to Arabic sentiment analysis has remained underexplored.

This study addresses this by integrating active learning into Arabic sentiment classification and evaluating its performance across three benchmark datasets that include both modern standard Arabic and dialectal content. Multiple deep learning models were employed under two annotation paradigms: Traditional human labeling and LLM-assisted annotation. Several state-of-the-art large language models were examined for their potential to accelerate the annotation process

The results demonstrate that active learning can significantly reduce annotation requirements without compromising classification accuracy. Moreover, LLM-assisted labeling proved to be an efficient and cost-effective complement to manual annotation, enhancing the overall scalability of sentiment analysis pipelines. These findings highlight the strong potential of combining active learning with LLM-driven annotation strategies to support sentiment analysis in low-resource languages such as Arabic.

Future work can proceed along three directions. First, extending experiments on additional datasets that span diverse Arabic dialects and domains (e.g., news, healthcare, education) would strengthen evidence of the framework’s generalizability. Second, employing LLMs fine-tuned on Arabic data for the labeling process could further improve annotation quality and adaptation to linguistic nuances. Third, exploring alternative sample selection strategies within the active learning loop, such as uncertainty-based, disagreement-based, or committee-based methods, and systematically comparing their effectiveness would provide deeper insights into optimizing labeling efficiency.

\bibliographystyle{elsarticle-num}
\bibliography{references}

\end{document}